# GIANT PANDA FACE RECOGNITION USING SMALL DATASET

*[a]Wojciech Michal Matkowski, [a]Adams Wai Kin Kong, [b]Han Su,*
*[c]Peng Chen, [c]Rong Hou,* and *[c]Zhihe Zhang*

[a] School of Computer Science and Engineering, Nanyang Technological University, Singapore, 639978
[b] College of Computer Science, Sichuan Normal University, 1819 Chenglong Road, Chengdu, Sichuan, China, 610101
[c] Chengdu Research Base of Giant Panda Breeding, Sichuan Key Laboratory of Conservation Biology for Endangered Wildlife, Chengdu, Sichuan, China, 610086

**ABSTRACT**

Giant panda (panda) is a highly endangered animal. Significant efforts and resources have been put on panda conservation. To measure effectiveness of conservation schemes, estimating its population size in wild is an important task. The current population estimation approaches, including capture-recapture, human visual identification and collection of DNA from hair or feces, are invasive, subjective, costly or even dangerous to the workers who perform these tasks in wild. Cameras have been widely installed in the regions where pandas live. It opens a new possibility for non-invasive image based panda recognition. Panda face recognition is naturally a small dataset problem, because of the number of pandas in the world and the number of qualified images captured by the cameras in each encounter. In this paper, a panda face recognition algorithm, which includes alignment, large feature set extraction and matching is proposed and evaluated on a dataset consisting of 163 images. The experimental results are encouraging.

*Index Terms—* biometrics, face recognition, giant panda, individual identification

## 1. INTRODUCTION

Recognition of individual animals is an important task, which allows researchers to understand animal's behavior and study population parameters such as population size, movement patterns, etc. There exist invasive and non-invasive methods for the individual recognition. Invasive methods require capturing an animal and physical tagging, which is troublesome, costly and likely to cause stress. Non-invasive methods use biometric traits such as DNA collected from hair or feces; or visual assessment based on images taken by cameras mounted in inhabited regions. DNA collection is costly and in some cases infeasible because it requires conservation workers going to rural or dangerous areas to acquire the samples. Therefore, visual assessment and animal biometrics gained attention as a potential solution.

Some animals such as zebra, tiger, whale shark and giraffe with unique patterns on their body are easy to be identified [1]. Fig. 1 shows different zebras and tigers. These patterns serve as their fingerprint. However, some animals are harder to be identified. Panda is one of them. To recognize animals for food safety (e.g., pig [2] and cow [3]), entertainment, sport (e.g., horse [4]) and wild animal conservation (e.g., chimpanzee [5]), biometric technology has been applied. More information about applying biometric technology to animals can be found in [1].

In the image processing community, researchers have studied panda. Chen et al. and Zhang et al. designed methods for panda face detection [7-9]. In addition to detection, Chen et al. attempted to estimate pose of panda in images [6]. Huang et al. [10] studied panda segmentation. According to our best knowledge, no one studied panda face recognition before.

Comparing with human face recognition and recognition of other animal species, panda face recognition is definitely a small dataset problem because according to the previous survey done in 2013 by National Forestry and Grassland Administration, China, it was estimated that there were only 1864 pandas in wild. In addition, in each encounter in wild, likely only several images with enough quality for recognition are available. Furthermore, pandas are generally solitary. In other words, in each small region, only several pandas can be captured by the same camera. Thus, in this paper, we study panda face recognition on a small dataset.

The rest of this paper is organized as follows. In Section 2, a panda face dataset is presented. In Section 3, the proposed algorithm is given in details. In Section 4, the experimental results are reported. In Section 5, the conclusion is given.

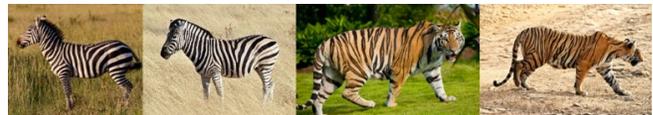
Fig. 1. Animals with unique patterns.

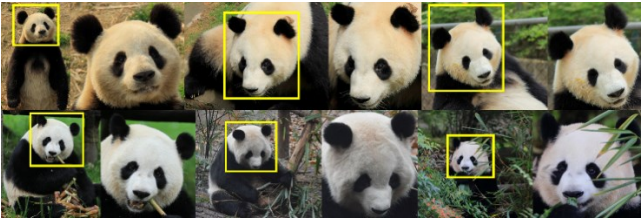

Fig. 2. Examples of original and cropped panda faces. Images in a row are from the same panda ID.

## 2. DATASET

Panda images used in this study were collected by Chengdu Research Base of Giant Panda Breeding. A dataset consists of 163 images of 28 different panda individuals (panda ID). Frontal face regions were manually cropped and resized to 100 pixels height, preserving the aspect ratio. Note that panda face detection is out of the scope in this study and there exist other works that focus on this task only. The panda face images have significant pose, illumination and occlusion variations. Pandas also have different facial expressions or often eat e.g., bamboo. Fig. 2 shows some examples of the original and cropped panda face images.

## 3. THE PROPOSED ALGORITHM

In this section, the proposed panda face recognition algorithm is presented. The algorithm consists of panda face alignment, feature extraction, and Partial Least Square (PLS) regression.

Panda images are usually captured in unconstrained environments. Thus, the panda face images have different poses, expression, illumination, shading, resolution and occlusion. To extract features in the same region of interest (ROI) from the panda face image, the alignment procedure illustrated in Fig. 3 and described in Algorithm 1 is used. First, Sobel edge detector is applied to image $I$ to find edges and get the key points. All the edge pixels' positions serve as 2D key points $k$. Coherent Point Drift (CPD) algorithm [11] is applied to find the correspondence between two sets of key points $k_t$ and $k_s$ and to estimate the affine transformation $\theta$ between target panda image $I_t$ and another panda image $I_s$. The subscript $t$ indicates target image, whereas the subscript $s$ indicates image to be wrapped. Wrapping is performed on image $I_s$ by using bicubic interpolation and the affine transformation $\theta$ calculated by CPD.

After image alignment, Local Binary Pattern (LBP) [12] and Gabor [13] histograms are extracted in each block from seven different grids (see Fig. 4) which cover panda face. These grids are designed to preserve spatial information and to capture it at different scales. Details of these grids, including the number of blocks in horizontal and vertical directions, are given in Table 1. Gabor orientation field is calculated as follows:

$$O(x,y) = \arg_{\theta_r} \max_{m,r} |G_{\lambda_{mk},\theta_r,\sigma_m,\gamma} * I|$$

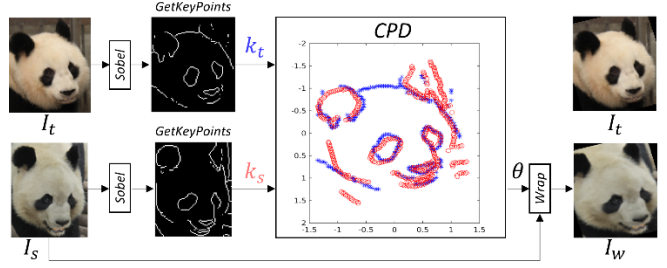

Fig. 3. Schematic illustration of wrapping image $I_s$ to image $I_t$. Wrapped image $I_s$ is denoted as $I_w$.

---
**Algorithm 1** Alignment
1: **procedure** ALIGN($I_s, k_t$)
2:    $k_s \leftarrow GetKeyPoints(Sobel(I_s))$
3:    $\theta \leftarrow CPD(k_s, k_t)$
4:    **return** $I_w \leftarrow Wrap(I_s, \theta)$
5: **end procedure**

---

where $G_{\lambda_{mr},\theta_r,\sigma_m\gamma}$ is a Gabor filter with an orientation $\theta_r = r\pi/8$, the wavelength of the sinusoidal component $\lambda_{mr}$, standard deviation of the elliptical Gaussian window $\sigma_m$, spatial aspect ratio $\gamma$, scale indices $m$ and orientation indices $r$. In the implementation, four scales and 16 orientations are used. Gabor orientation fields are calculated over the grayscale image and their histograms extracted in each block are concatenated to form Gabor features $h_{Gabor}$. The Local Binary Pattern $LBP_{P,R}$ is calculated as follows:

$$LBP_{P,R} = \sum_{p=0}^{P-1} s(g_p - g_c) 2^p$$

where $P$ is the number of sampling points, $R$ is the radius of the sampling circle, $g_c$ is the center pixel value, $g_p$ is the neighbourhood pixel value and $s(a) = 1$ if $0 \leq a$, otherwise $s(a) = 0$. In the implementation $LBP_{8,1}^{riu2}$ and $LBP_{8,2}^{u2}$ are used. $LBP_{8,1}^{riu2}$ and $LBP_{8,2}^{u2}$ are calculated over R, G and B channels separately and their histograms extracted in the blocks as indicated in Table 1 are concatenated to form the LBP features $h_{LBP}$. LBP features $h_{LBP}$ and Gabor features $h_{Gabor}$ are concatenated to make a large 15374 dimensional feature vector $x$ (see Algorithm 2).

Table 1. Details of seven different grids.

| Grid | Hor. dir. | Ver. dir. | No. blocks | Features extracted |
|---|---|---|---|---|
| G1 | 7 | 5 | 35 | $LBP_{8,1}^{riu2}$, Gabor |
| G2 | 5 | 7 | 35 | $LBP_{8,1}^{riu2}$, Gabor |
| G3 | 5 | 5 | 25 | $LBP_{8,2}^{u2}$, Gabor |
| G4 | 4 | 3 | 12 | $LBP_{8,2}^{u2}$, Gabor |
| G5 | 3 | 4 | 12 | $LBP_{8,2}^{u2}$, Gabor |
| G6 | 3 | 3 | 9 | $LBP_{8,2}^{u2}$, Gabor |
| G7 | 2 | 2 | 4 | $LBP_{8,2}^{u2}$, Gabor |

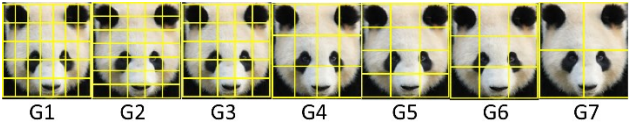

G1　G2　G3　G4　G5　G6　G7

Fig. 4. Illustration of seven different grids.

---

**Algorithm 2** Feature Extraction
1: **procedure** EXTFEAT($I$)
2:   $h_{LBP} \leftarrow ExtLBP(I)$
3:   $h_{Gabor} \leftarrow ExtGabor(rgb2gray(I))$
4:   **return** $x \leftarrow [h_{LBP}, h_{Gabor}]$
5: **end procedure**

---

The proposed algorithm consists of enrolment and testing phases. During enrolment, the classifiers $\beta$ are built, whereas during testing, these classifiers are used to calculate the comparison scores between panda images and perform the recognition. Both procedures include an image to image alignment and feature extraction. The classifiers $\beta$ are Partial Least Square (PLS) regression coefficients calculated using NIPALS algorithm whose details can be found in [14]. PLS has proven to handle high dimensional feature vectors and imbalanced datasets well [15].

The enrollment is illustrated in Fig. 5. In the enrollment procedure, before training the PLS classifier $\beta_t$ for one target panda image $I_t$, all other training images are aligned to the target image and the features are extracted. The PLS classifier is used in one-against-all approach, which means that images coming from the same panda ID $c$ are labelled as $y_i = 1$, otherwise $y_i = -1$. Classifiers are built in this manner for each panda image $I_t$ giving totally $N$ classifiers stored in a classifier matrix $\beta = [\beta_1, ..., \beta_N]$. The pseudocode of the enrollment procedure is given in Algorithm 3. Note that $X$ and $y$ in Algorithm 3 are standardized (z-score) before feeding into PLS. Additionally, the target key points $k = [k_1, ..., k_N]$ are also stored. In the notation $\beta_t$, $x_i$ are row vectors and $y_i$ is a scalar.

---

**Algorithm 3** Enrollment
1: **procedure** ENROLL($I, I_t, c$)
2:   $k_t \leftarrow GetKeyPoints(Sobel(I_t))$
3:   **for** $i \leftarrow 1\ to\ N$
4:     $I_i \leftarrow$ ALIGN($I_i, k_t$)
5:     $x_i \leftarrow$ EXTFEAT($I_i$)
6:     **if** $GetPandaID(I_i) = c$ **then**
7:       $y_i \leftarrow 1$
8:     **else**
9:       $y_i \leftarrow -1$
10:     **end if**
11:   **end for**
12:   $X = [x_1, ..., x_N]$
13:   $y = [y_1, ..., y_N]$
14:   **return** $\beta_t \leftarrow PLS(X, y), k_t$
15: **end procedure**

---

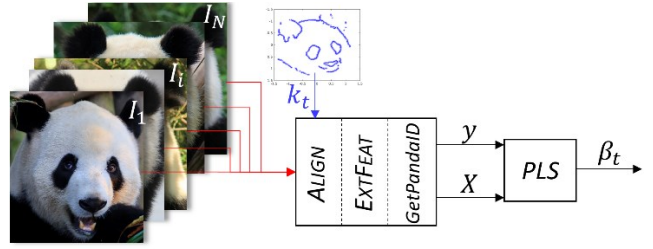

Fig. 5. Schematic illustration of the enrollment procedure, where $k_t$ are the key ponints from the target panda image $I_t$ and $\beta_t$ is the trained classifier.

In the testing procedure, a probe image $I_q$ is aligned to each gallery image and the features are extracted. More precisely, the image is aligned by using the key points $k$, which are obtained in the enrollment and $N$ feature vectors are extracted from the wrapped images to form a feature matrix $X$. Matching is performed by $X\beta^T$, where $T$ is a matrix transpose. The comparison scores $\hat{y}$ are obtained by taking the diagonal values of the resultant matrix. The maximum score of each panda ID classifier is the final prediction. The testing procedure is given in Algorithm 4.

---

**Algorithm 4** Test
1: **procedure** TEST($I_q, \beta, k$)
2:   **for** $i \leftarrow 1\ to\ N$
3:     $I_i \leftarrow$ ALIGN($I_q, k_i$)
4:     $x_i \leftarrow$ EXTFEAT($I_i$)
5:   **end for**
6:   $X = [x_1, ..., x_N]$
7:   **return** $\hat{y} \leftarrow diag(X\beta^T)$
8: **end procedure**

---

## 4. EXPERIMENTAL RESULTS

The proposed algorithm is evaluated using the leave-one-out approach because the dataset is very small. In particular, in the validation, each probe image panda ID has always a corresponding panda ID enrolled in the gallery (closed set) and the gallery does not contain the probe image. It means that each probe image is excluded from the training and then it is evaluated on 162 classifiers.

The evaluation is performed on verification and identification tasks. In the verification, a probe image is compared with another panda ID enrolled in the gallery, in order to determine whether the probe image ID and the gallery image ID are the same (1 to 1 comparison). In the identification, a probe image is compared with all images enrolled in the gallery (1 to N comparisons), in order to determine the ID of the probe image. As the evaluation metric of verification accuracy, True Acceptance Rates (TAR) at 1% False Acceptance Rates (FAR) are used, and for identification, rank-1 accuracy are reported.

Table 2. Verification accuracy TAR (%) at 1% FAR and
Rank-1 identification accuracy (%).

| Method | @ 1% FAR | Rank-1 |
|---|---|---|
| Proposed | **65.64** | **71.17** |
| ResNet-50 [18] | 59.51 | 62.58 |
| VGG-16 [19] | 55.21 | 57.06 |
| AlexNet [16] | 49.01 | 52.15 |
| GoogLeNet [17] | 54.60 | 60.74 |
| VGG-Face [20] | 50.92 | 54.60 |

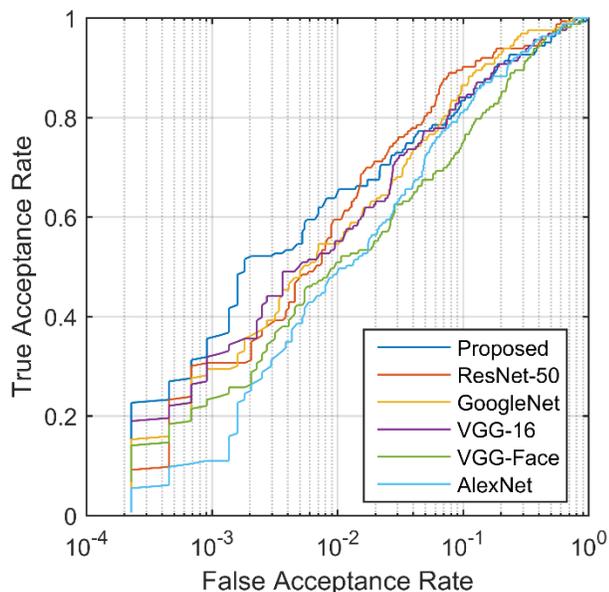

Fig. 6. ROC curves of the proposed algorithm and five deep learning methods.

Five state-of-the-art deep learning architectures AlexNet [16], GoogLeNet [17], ResNet-50 [18], VGG-16 [19] and VGG-Face [20] are trained for comparison. All these networks are first pretrained on the ImageNet dataset, except the VGG-Face, which is pretrained on human face images dataset [20]. Before fine-tuning, the input panda face images are resized to fit the network input requirements. These deep architectures require large datasets for training, and thus in order to minimize the risk of overfitting, only the last softmax layer is fine-tuned on this small panda image dataset. The performance of the proposed algorithm and the deep learning based methods are reported in Table 2. Additionally, the Receiver Operating Characteristic (ROC) curve is presented in Fig. 6. Examples of true and false acceptance cases of the proposed algorithm are shown in Figs. 7 and 8, respectively. The results show that the proposed algorithm outperforms all five competing deep learning based methods on the small panda image dataset. It achieves 6.43% and 8.59% higher accuracy than the second best ResNet-50 for the verification and identification tasks, respectively.

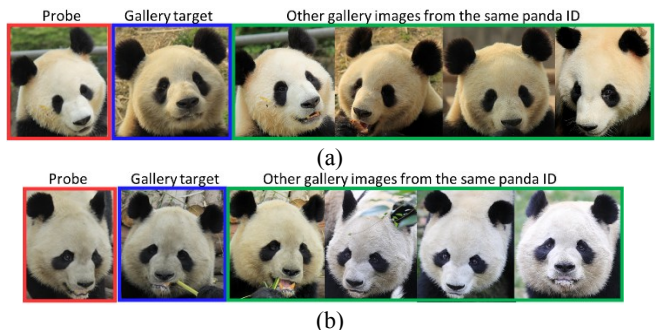

Fig. 7. Examples of true accepted pandas at 1% FAR. Probe images are in red boxes. Gallery target images are in blue boxes. Four gallery images from the same panda ID, which are also used to build the classifier for the gallery target panda image are in green boxes.

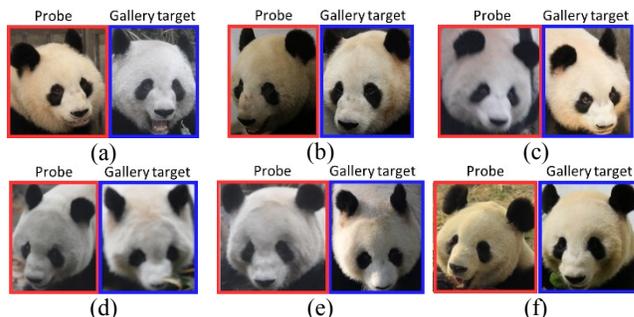

Fig. 8. Examples of false accepted pandas at 1% FAR. Probe images are in red boxes. Gallery target images are in blue boxes.

## 5. CONCLUSION

Recognizing individual animals is important for food safety, sport and animal conservation. Though many animals have been studied, panda, the logo of the World Wildlife Fund, was neglected. In this paper, we present the first study on panda face recognition. Although pandas look very similar and even the staff of Chengdu Research Base of Giant Panda Breeding have difficulties in recognizing them, the experimental results show that panda faces still have some discriminative features to the proposed algorithm. These results are encouraging. In the coming future, we will improve the proposed algorithm by using the fined details from facial components, such as eyes and ears and collect more data for algorithm development and evaluation.

## 6. ACKNOWLEDGMENTS

This research is supported by the National Natural Science Foundation of China (31300306), the Sichuan Science and Technology Program (2018JY0096), Chengdu Research Base of Giant Panda Breeding (NO. CPB2018-02), the Chengdu Giant Panda Breeding Research Foundation (CPF Research 2014-02, 2014-05), and the Panda International Foundation of the National Forestry Administration, China (CM1422, AD1417). The research done in the Nanyang Technological University, Singapore is under the project, Using Image Analysis for Giant Panda Verification.